\documentclass[a4paper, twocolumn, DIV25,9pt]{scrartcl}
\pdfoutput=1
\usepackage{palatino}
\usepackage{compactbib}

\usepackage{graphicx}
\usepackage{natbib}
\usepackage{compactbib}
\usepackage{fixltx2e}
\usepackage{float}
\usepackage{url}

\date{}

\newtheorem{definition}{Definition}

\title{\flushleft{Crossing the Reality Gap: a Short Introduction to the Transferability Approach}\\~\\
\normalsize Jean-Baptiste Mouret, Sylvain Koos and St\'ephane Doncieux\\
Universit\'e Pierre et Marie Curie-Paris 6, France -- CNRS UMR 7222, France\\ mouret@isir.upmc.fr
}
\author{}
\begin{document}
\maketitle

\begin{abstract}
\bfseries
In robotics, gradient-free optimization algorithms (e.g. evolutionary
algorithms) are often used only in simulation because they require the
evaluation of many candidate solutions. Nevertheless, solutions
obtained in simulation often do not work well on the real device. The
transferability approach aims at crossing this gap between simulation
and reality by \emph{making the optimization algorithm aware of the
  limits of the simulation}.

In the present paper, we first describe the transferability function,
that maps solution descriptors to a score representing how well a
simulator matches the reality. We then show that this function can be
learned using a regression algorithm and a few experiments with the
real devices. Our results are supported by an extensive study of the
reality gap for a simple quadruped robot whose control parameters are
optimized. In particular, we mapped the whole search space in reality
and in simulation to understand the differences between the fitness
landscapes.
\end{abstract}
\normalfont

\section{Introduction}
\begin{figure*}[htb!]
  \begin{minipage}{0.5\linewidth}
    \begin{center}
      \includegraphics[height=5cm]{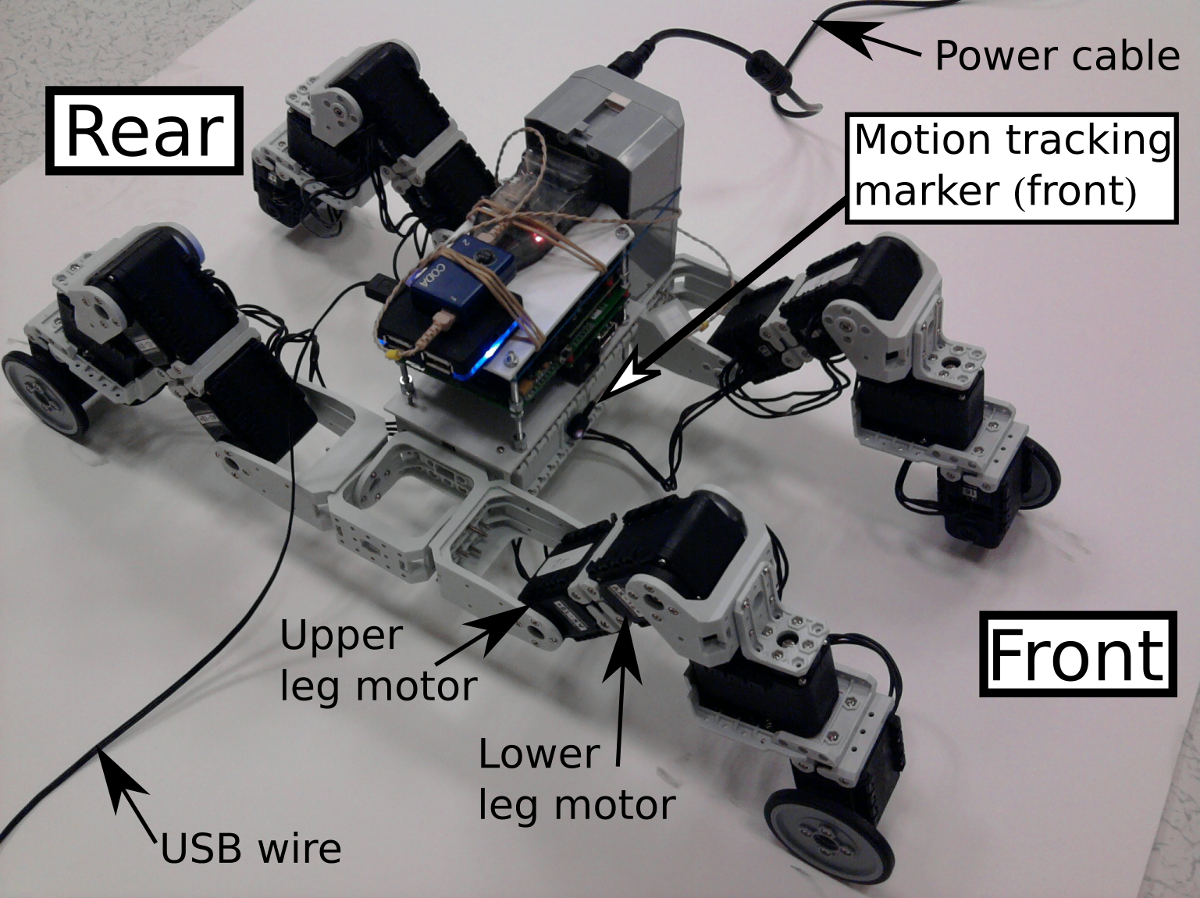}\\
      (a)
    \end{center}
  \end{minipage}
  \begin{minipage}{0.5\linewidth}
    \begin{center}
      \includegraphics[height=5cm]{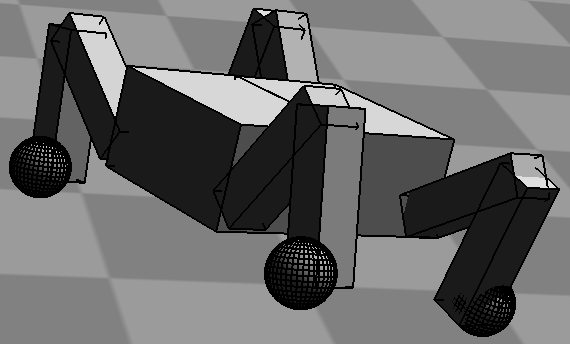}\\
      (b)
    \end{center}
  \end{minipage}
\caption{\label{fig:robot}(a) The quadruped robot is based on a
  Bioloid robotic kit (Robotis). It has 8 degrees of freedom (wheels
  are not used in this experiment). We track its movement using a 3D
  motion capture system. (b) The simulation is based on the
  Bullet dynamic simulator \citep{Boeing2007}.}
\end{figure*}

Gradient-free optimization algorithms underlie many machine learning
methods, from policy search techniques~\citep{Whiteson2006,
  Heidrich-Meisner2008a} to automatic design
approaches~\citep{Lohn2006, Lipson2000}. They are also one of the tool
of choice for researches in embodied intelligence because they make
possible to obtain artifacts (e.g. neural networks to control a robot)
without having to describe their inner workings~\citep{Pfeifer2006}.

In many of their applications, these algorithms spend most of their
running time in evaluating the quality of thousands of potential
solutions. This observation encourages many researchers to work with
simulations instead of real devices, because simulations are usually
cheaper and faster than real experiments. For instance, most published
work in evolutionary robotics (ER) --- in which researchers typically
aim at finding original controllers for robots --- is carried with
simulated robots~\citep{Doncieux2011,Nelson2009}. At first sight,
robots are articulated rigid bodies for which many simulations tools
exist; it is therefore tempting to suppose that an efficient result
obtained by optimizing in simulation will work similarly on the real
robot. Unfortunately, no simulator is perfect and optimization
algorithms have no reason to avoid exploiting every badly modeled
phenomena that increase performance. It is thus often observed that
solutions optimized in simulation are inefficient on the real
robot. On the contrary, most engineers intuitively know the limit of
their simulation tools and avoid relying on what is incorrectly
modeled.

This difference in performance with a simulation and with the real
device has been termed the \emph{reality gap}. It is of course not
restricted to ER since the same issues are encountered with all the
other optimization algorithms and with many other experimental
setups. However, we will restrict our current discussion to ER because
the reality gap is central in this community. The reality gap is
indeed arguably one of the main issue that prevent a widespread use of
evolutionary algorithms to optimize parameters of robot controllers:
evaluating every potential solutions in reality is very costly because
it requires complex experimental setups and a lot of time; evaluating
potential solutions in simulation is cheaper and faster but it often
leads to solutions that cannot be used on the real device. How could we
proceed?

If we do not reject the use of simulators, the first idea to reduce
this gap is to design better simulators. Such an approach can work up
to a certain extent but complex simulators are slow (e.g. simulating
fluids can be slower than reality) and even the best simulators cannot
be infinitely exact. An attractive idea is to automatically design a
simulator, for instance by learning a surrogate model of the fitness
function~\citep{Jin2005b}, or, following a related idea, to automatically
improve an existing
simulator~\citep{Bongard2006,Zagal2007}. Nevertheless, creating an
algorithm that automatically designs the perfect simulator appears at
least as difficult as designing evolutionary algorithms to learn the
optimal behaviors of a robot. Moreover, these methods will never
accurately model every possible force that can act on a device. For
instance, it is hard to expect that an algorithm will automatically
discover a good model of fluid dynamics in a classic rigid-body
simulator, whatever the improvements of the simulator are.

As an alternative approach, \cite{Jakobi1997} proposed to prevent the
optimization algorithm to exploit badly modeled phenomena by hiding
them in an ``envelope of noise''. Despite some success with the
evolution of a controller for an hexapod robot, Jakobi did not
describe any generic method to choose what has to be noised and how
this noise should be applied. Applying the ``envelope of noise''
technique therefore often requires a lot of experiments and fine-tuning
of the simulator, which is exactly what researchers try to avoid when
designing optimization algorithms. For instance, it is hard to know
how to add noise when evolving locomotion controllers for legged
robots.

In the present paper, we describe a recently introduced approach to
cross the reality gap: the transferability
approach \citep{Koos2011}. Our aim is to give a didactic presentation
of the intuitions that guide this method as well as the main results
obtained so far. The interested reader can refer to \citep{Koos2011}
for detailed results and discussions.

\section{Experimental Apparatus}
\paragraph{Robot and controller}
We studied the reality gap with an 8-DOFs quadruped robot made from a
Bioloid Kit (Fig.~\ref{fig:robot}). Another experiment inspired by
Jakobi's T-maze is reported in \citep{Koos2011}.

The angular position of each motor follows a sinusoid. All these
sinusoidal controllers depend on the same two real parameters
$(p_1,\ p_2) \in [0, 1]^2$ as follows:

\begin{displaymath}
\alpha(i, t) = \frac{5 \pi}{12} \cdot dir(i) \cdot p_1 - \frac{5
  \pi}{12} \cdot p_2 \cdot \sin(2 \pi t - \phi(i))
\end{displaymath}

where $\alpha$ denotes the desired angular position of the motor $i$
at time-step $t$. $dir(i)$ is equals to 1 for both motors of the
front-right leg and for both motors of the rear-left leg; $dir(i)= -1$
otherwise (see Fig. \ref{fig:robot} for orientation). The phase angle
$\phi(i)$ is 0 for the upper leg motors of each leg and $\pi/2$ for
the lower leg motors of each leg. Both motors of one leg consequently
have the same control signal with different phases. Angular positions
of the actuators are constrained in $[-\frac{5 \pi}{12}, \frac{5
    \pi}{12}]$.

The fitness is the distance covered by the robot in 10\linebreak seconds.

\paragraph{Reality gap}
We first followed a typical ER approach: we evolved controllers in
simulation and then transferred the best solution on the robot. On
average (10 runs), the best solution in simulation covered 1294 mm
(sd = 55mm) whereas the same controller leads to only 411 mm in
reality (sd = 425mm); thus we observe a clear reality gap in this task.

This reality gap mostly stems from classic issues with dynamic
simulations of legged robots. In particular, contact models are not
accurate enough to finely simulate slippage, therefore any behavior
that relies on non-trivial contacts will be different in reality and in
simulation. Dynamical gaits (i.e. behaviors for which the robot is
often in unstable states) are also harder to accurately simulate than
more static gaits because the more unstable a system is, the more
sensitive it is to small inaccuracies.

\begin{figure*}[htb!]
  \begin{minipage}{0.5\textwidth}
    \includegraphics[width=\textwidth]{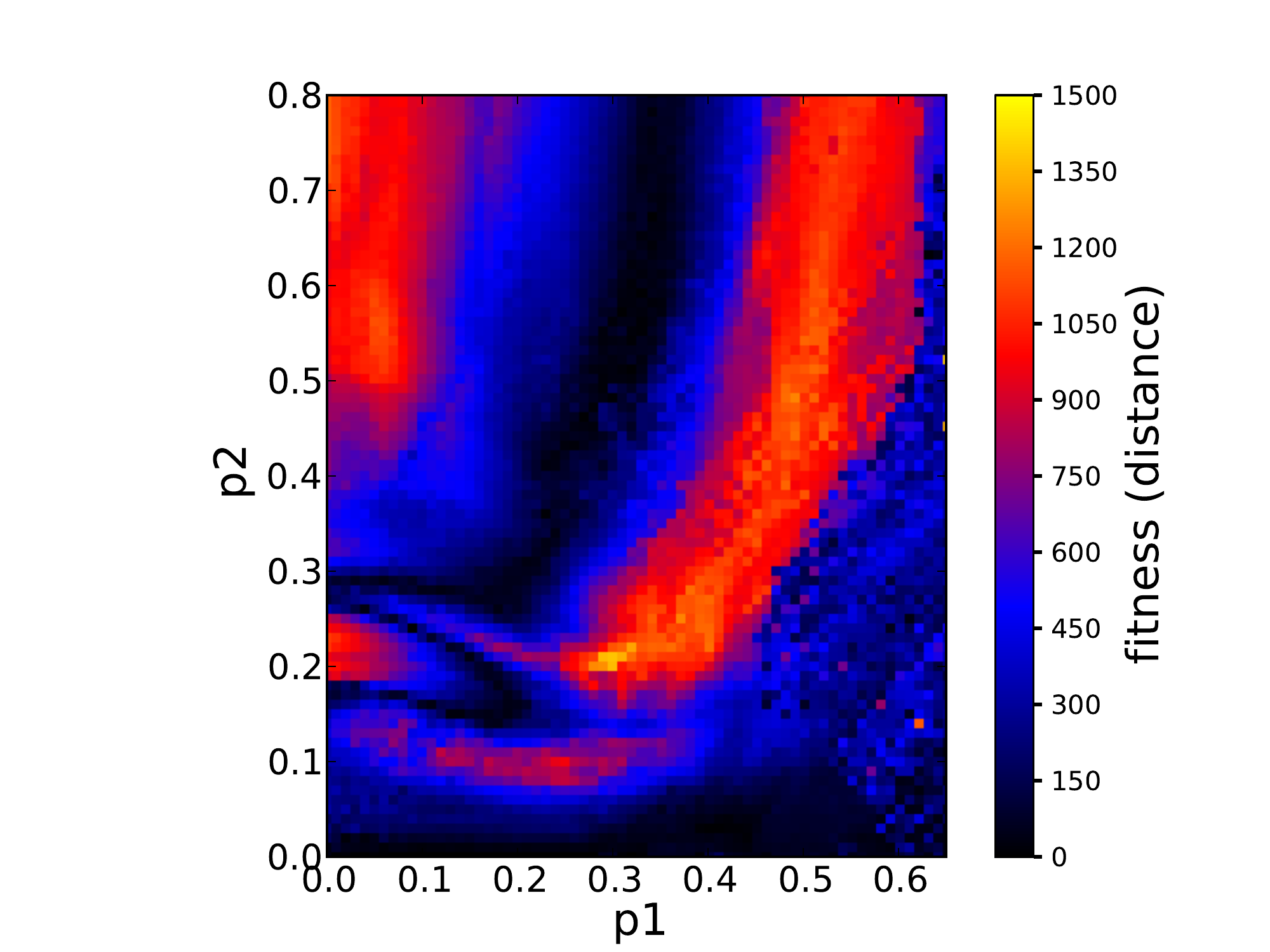}
  \end{minipage}
  \begin{minipage}{0.5\textwidth}
    \includegraphics[width=\textwidth]{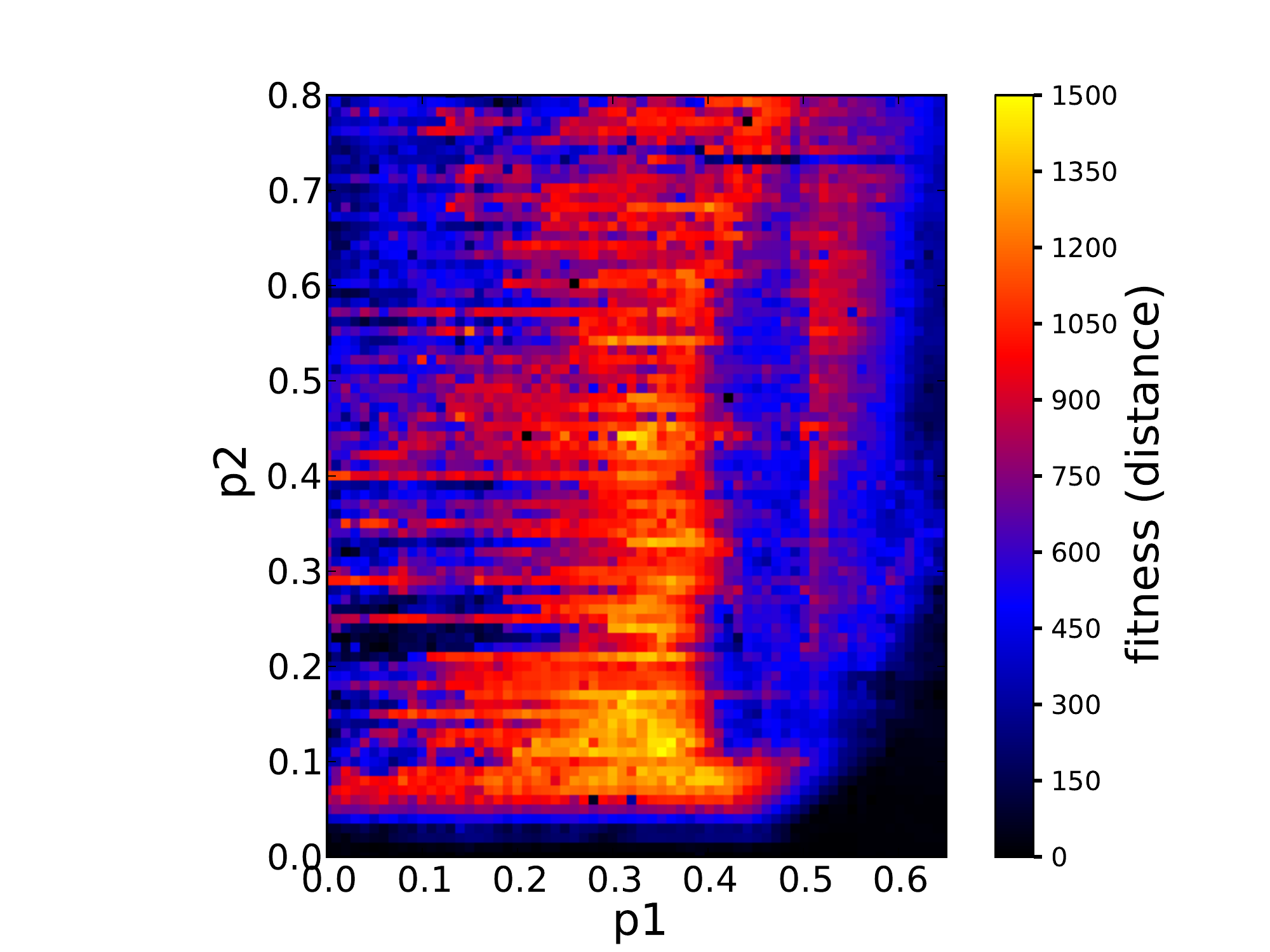}
  \end{minipage}  
  \caption{\label{fig:landscape}(left) Fitness landscape in the dynamic simulator (5500
    experiments). (right) Fitness landscape with the real robot (5500
    experiments). In both maps, $p_1$ and $p_2$ are the evolved parameters of the controller.}
\end{figure*}

The small number of parameters of this controller allows the mapping
of the whole search space. We realized 5500 experiments on the real
robot and interpolated the rest of the search space
(Fig.\ref{fig:landscape}(a)). We also mapped the fitness landscape in
simulation (5500 experiments, Fig.\ref{fig:landscape}(b)). To our
knowledge, this is the first time that we are able to visualize a
fitness landscape for a real robot and its simulation.

The differences between the two landscapes correspond to the reality
gap. The landscape in simulation contains four main fitness peaks and
one global optimum. The landscape obtained in reality is noisier but
simpler and it seems to contain only one important fitness peak. In
both landscapes, we observe a large low-fitness zone but the main
high-fitness zones match only a small zone and for only one fitness
peak. A typical example of reality gap will occur if the optimization
in simulation leads to solutions in the top left corner of the
fitness landscape, for which solutions have a high fitness in
simulation but a very bad one in reality.

\begin{figure}
  \begin{minipage}{0.5\textwidth}
    \includegraphics[width=\textwidth]{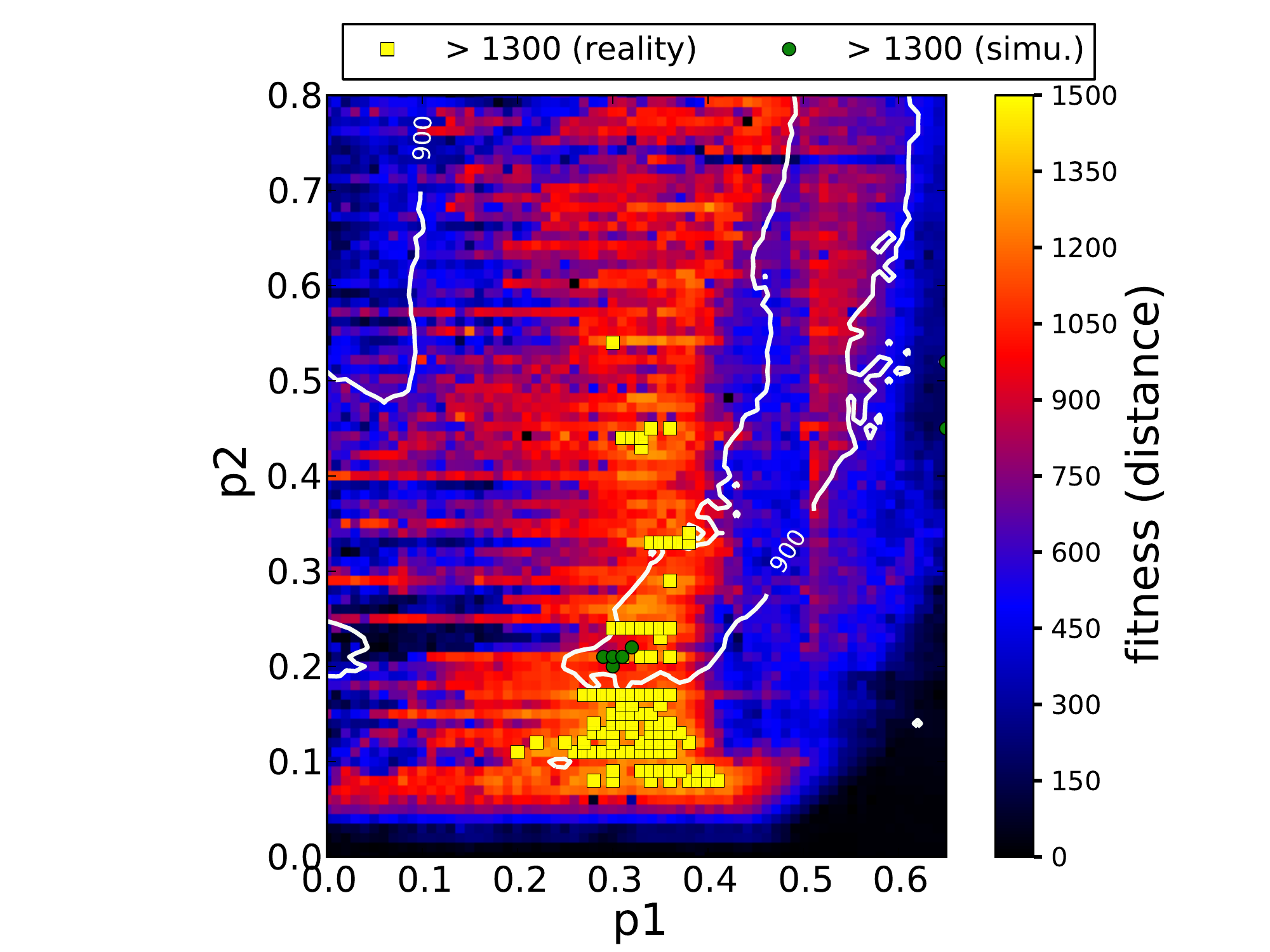}
  \end{minipage}
  \caption{\label{fig:landscape900}Superposition of the fitness landscape in reality with the one obtained in simulation. The contour line denotes the zones for which the simulation leads to high fitness values (greater than 900mm).}
\end{figure}

While the fitness landscapes in simulation and in reality are very
different, there exist a lot of controllers with a good fitness
(greater than $900$ mm; these controllers achieve gaits comparable to
those obtained with hand-tuned controllers) in both simulation and
reality (Fig.\ref{fig:landscape900}). If we visually compare gaits
that correspond to this zone in simulation and in reality, we observe
a good match.

Our interpretation is that the simulation is accurate in at least this
sub-part of the search space

\begin{figure*}
  \begin{minipage}{0.5\textwidth}
    \begin{center}
      \includegraphics[width=\linewidth]{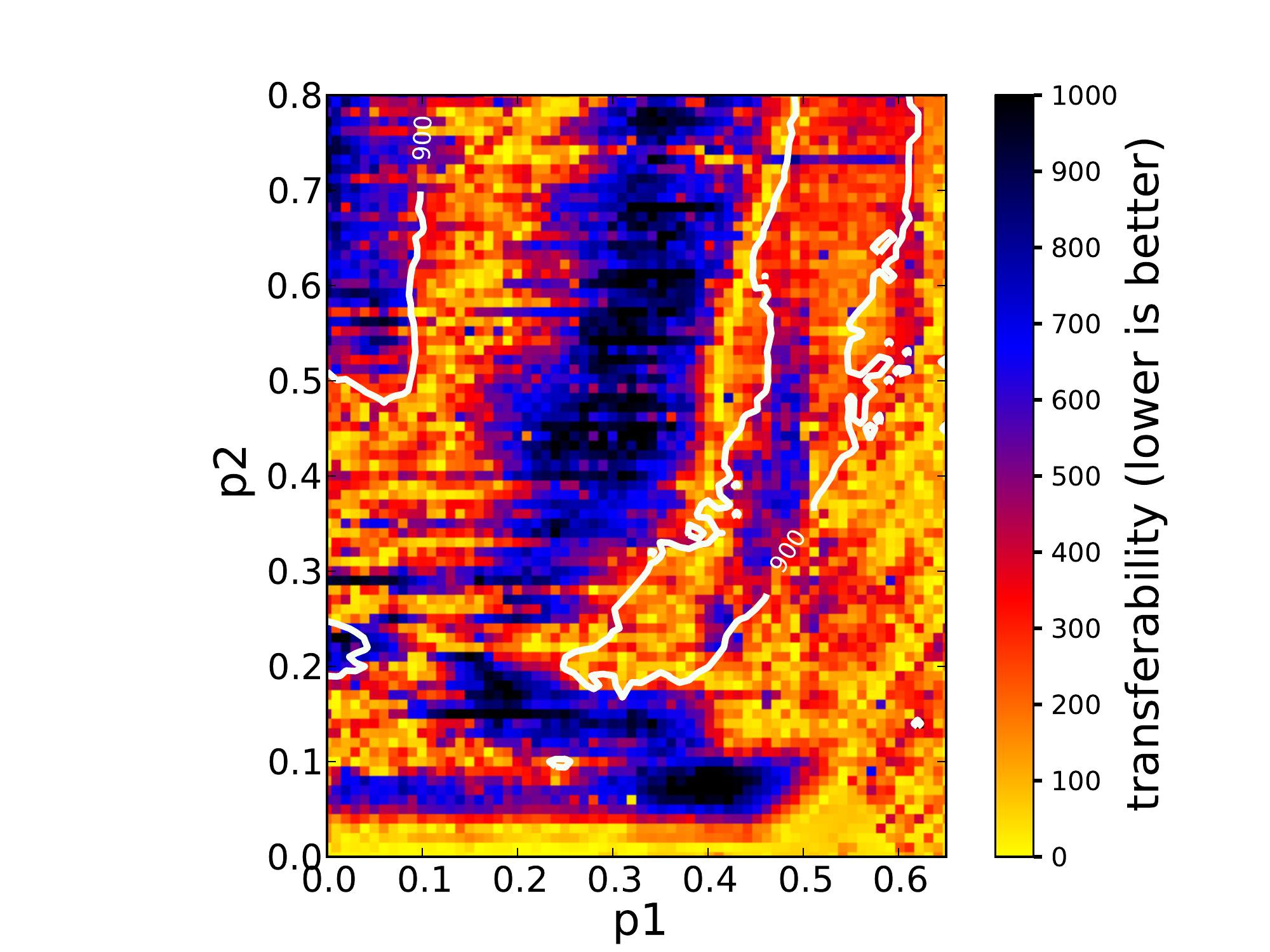}\\
    \end{center}
\end{minipage}
  \begin{minipage}{0.5\textwidth}
    \begin{center}
      \includegraphics[width=\linewidth]{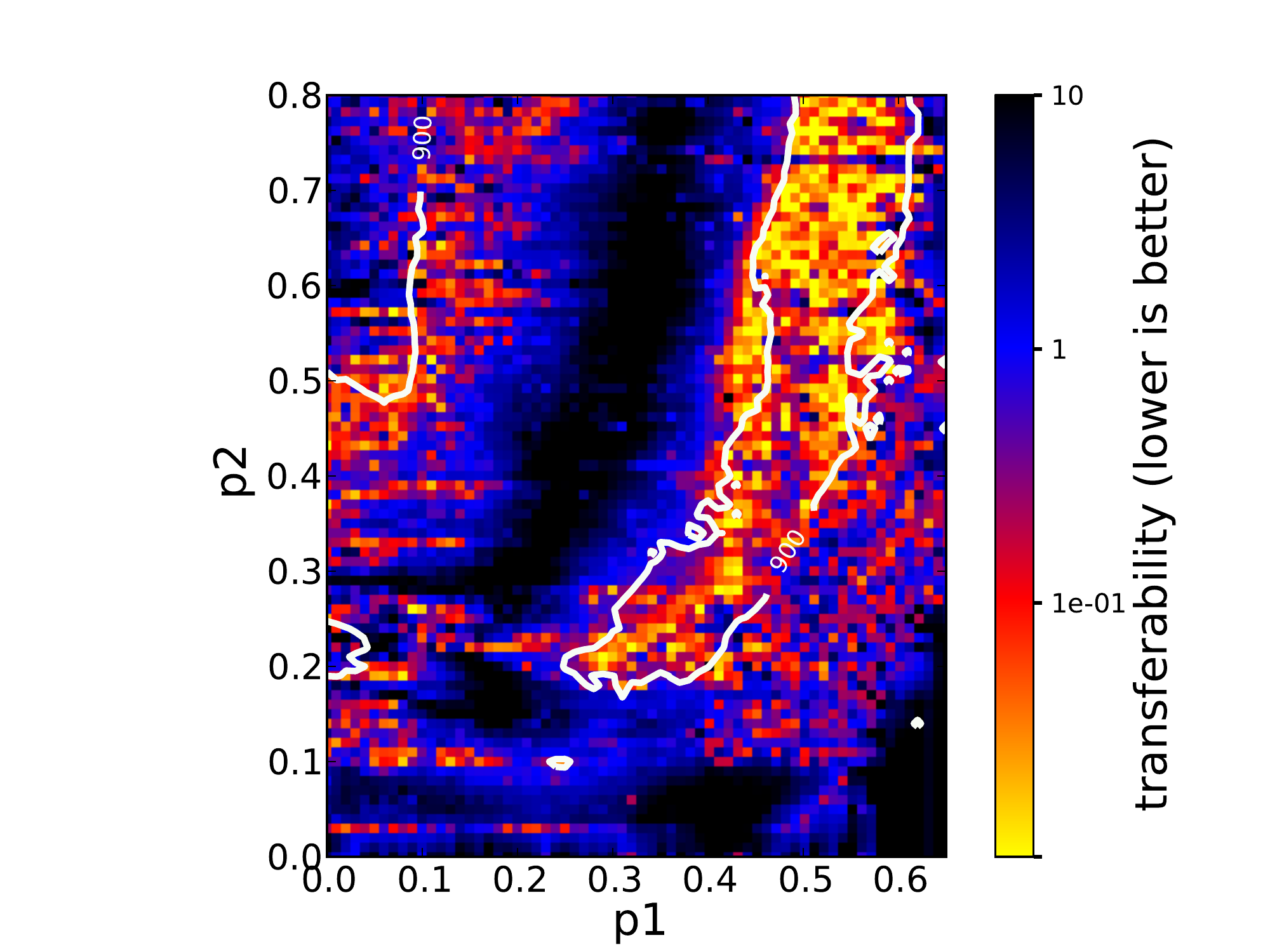}\\
    \end{center}
  \end{minipage}
\caption{\label{fig:transf}(left) Transferability function based on the difference of fitness values. (right) Transferability function based on the difference of trajectories. In both maps, the contour line denotes the zones for which the simulation leads to high fitness values (greater than 900mm).}
\end{figure*}

\section{The Transferability Function}
This interpretation leads to the fundamental hypothesis of the
transferability approach: \emph{for many physical systems, it is
  possible to design simulators that work accurately for a subset of
  the possible solutions}. In the case of dynamic simulators,
physicists work on the dynamic of rigid body since the XVII-th century
and the accumulated knowledge allows engineers to make good
predictions for many physical systems.

Since simulations will never be perfect, our approach is to \emph{make
  the system aware of its own limits} and hence allows it to avoid
solutions that it cannot correctly simulate. These limits can be
captured by a transferability function:

\begin{definition}[Transferability function]
  A \emph{transferability function} is a function that maps, for the
  whole search space, descriptors of solutions (e.g. genotypes,
  phenotypes or behavior descriptors) to a \emph{transferability
    score} that represents how well the simulation matches the
  reality.
\end{definition}

There are many ways to define a similarity, therefore there are many
possible transferability functions. Describing the similarity of
behaviors in robotics has recently been investigated in the context of
diversity preservation~\citep{Mouret2012} and novelty
search~\citep{Lehman2010}. Many measures have been proposed. For a
legged robot, one can compare covered distance (i.e. compare the
fitness values), trajectory of the center of mass at each time-step,
angular positions of each joint for each time-step, contact of the
legs with the ground, ...  At any rates, the best similarity measure
highly depends on the task and on the simulator.

The most intuitive input space for the transferability function is the
genotype space. However, the maps from genotype to transferability may
be very non-linear because the relationship between genotypes and
behaviors is often complex in evolutionary robotics. Many genotypes
(e.g. neural networks or development programs) are also hard to put as
the input of functions. An alternative is to use the behavior in
simulation, which is easy to obtain. The transferability function then
answers the question: ``given this behavior in simulation, should we
expect a similar behavior in reality?''. For instance, in many dynamic
simulations we observe robots that unrealistically jump above the
ground when they hit it. If the 3D-trajectory of the center of mass is
used as an input space, then the transferability function will easily
detect that if the z-coordinate is above a threshold, then the
corresponding behavior is not transferable at all.

For the considered quadruped robot, we computed two transferability
functions:
\begin{itemize}
\item input space: genotype; similarity measure:difference in covered distance (fitness) (Fig.\ref{fig:transf}(a));
\item input space: genotype; similarity measure: sum of the squared Euclidean distance between each point of the 3D trajectories of the geometrical center of the robot (Fig.\ref{fig:transf}(b)).
\end{itemize}
 In both cases
we observe that the high-fitness zone of the simulation in the top
left corner is not transferable but a large part of the solutions from
the other high fitness zone appears transferable.

\section{Learning the Transferability Function}
For evolutionary robotics, it is obviously unfeasible to compute the
transferability score for each solution of the search space -- as we
did it in these simple experiments -- because this would require to
test every point of the search space on the real robot. To avoid this
issue, \emph{the main proposition of the transferability approach is
  to automatically learn the transferability function using supervised
  learning techniques}. Using a few tests on the real system and a few
evaluations in simulation, we propose to use a regression technique
(e.g. a neural network or a support vector machine) to predict how
well simulation and reality will match for any solution of the search
space. This predictor will thus estimate the transferability of each
potential solution. Put differently, the transferability approach
proposes to learn the limits of the simulation.

It may seem counter-intuitive and inefficient to approximate the
transferability instead of the fitness (i.e. using a surrogate model
of the fitness), but working with the transferability function is
promising for at least two reasons. First, approximating the fitness
function for a dynamic system (e.g. a robot) means using a few tests
on the real robot to build the whole fitness landscape. In the same
way as simulators will never be perfect, this approximation will not be
perfect, therefore we will likely face reality gap issues. Second,
learning the fitness function is likely to be harder than learning the
transferability function. Indeed, using a machine learning technique
to learn the fitness function of a robot is equivalent to
automatically design a simulator for a complex robot: the function has
to predict a description of the behavior (the fitness) from a
description of the solution (the genotype). Such a simulator would
therefore need to include the laws of articulated rigid body dynamics,
but these laws are unlikely to be correctly discovered using a few
trajectories of a robot. On the contrary, predicting that a solution
will not be transferable can often be done using a few simple criteria
that a machine learning algorithm can find. For instance, a
classification algorithm could easily predict that high-frequency
gaits are not transferable by applying a threshold on a frequency
parameter (the ease of prediction depends on the input space of the
predictor). In summary, learning the transferability
\emph{complements} a state-of-the art simulator instead of reinventing
or improving it.

\section{Finding Efficient and Transferable Solutions}
Using a simulator to find solutions that perform well in reality can
be restated as a two-objective optimization problem, where the
objectives are (1) the performance in simulation and (2) the accuracy of
the simulation for the tested solution. Optimal solutions for this
problem will be perfectly simulated and perfectly efficient in
simulation. However, there is no reason to believe that the best
solutions in simulation will correspond to the best solutions in
reality. On the contrary, the best solutions in simulation are often
highly dynamic behaviors that strongly rely on unrealistic effects; the
best solutions in reality will also be probably highly tuned behaviors
instead of simpler, more robust behaviors.

We therefore expect to see a trade-off between transferability and
fitness in simulation. Multi-objective evolutionary algorithms (MOEA,
see ~\cite{Deb2001}) are well suited methods for this two-objective
optimization:
\begin{displaymath}
  \textrm{maximize } \left\{\begin{array}{l}
  \textrm{fitness}(x)\\
  \textrm{approximated transferability}(x)
  \end{array}
  \right.
\end{displaymath}
Nonetheless, we are essentially optimizing the fitness under the
constraint of the transferability. While MOEAs are recognized tools to
apply soft constraints~\citep{Fonseca1998}, other constrained optimization
algorithms could also be employed.

\begin{figure}
  \begin{center}
    \includegraphics[width=0.75\linewidth]{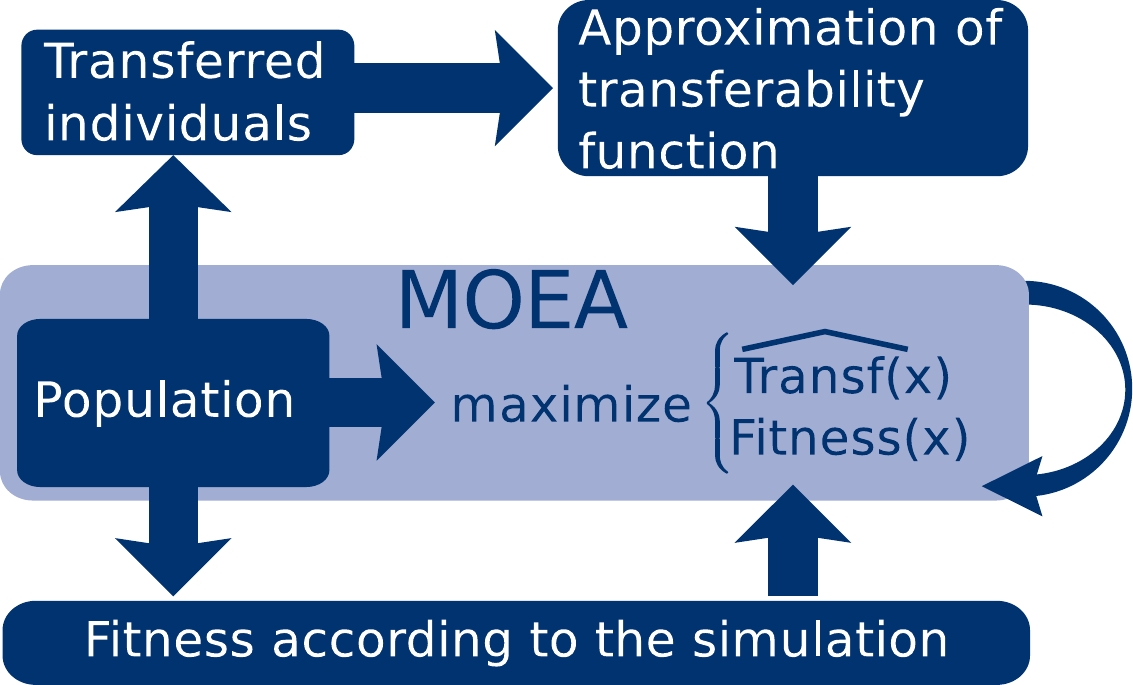}
  \end{center}
  \caption{\label{fig:principle}Principle of the multi-objective
    optimization of both the fitness and the
    transferability. Individuals from the population are periodically
    transfered on the robot to improve the approximation of the
    transferability function.}
\end{figure}

We chose to use the Inverse Distance Weighting (IDW) method to
approximate the transferability function because it's simple and
efficient enough. This method can be substituted with any other
regression/interpolation method.

An interesting question is \emph{when} to improve the approximation,
that is when to transfer an individual to evaluate it in reality. A
first option is to transfer solutions before launching the
optimization, build the approximation and do not modify it during the
optimization. Another option is to transfer a few individuals before
the first generation, in order to initiate the process, and then
periodically update the approximation by transferring one of the
candidate solution of the population. The second option has the
advantage of focusing the approximation on useful candidate solutions
because the population will move towards peaks of high fitness. While
the first option is simpler, we chose the second one in our current
implementation: every 20 generations, the individual from the
population that is the most different from the others is tested on the
real robot. At the end of the optimization, we select the solution
with the best fitness and above a user-defined value for the
transferability.

Figure \ref{fig:principle} summarizes this process. Our source code is
available on EvoRob\_db (\url{http://www.isir.fr/evorob_db}).

\section{Experimental Results}
The two objectives are optimized with the NSGA-II algorithm because
it's a classic and versatile MOEA. The size of the population is 40
and the algorithm is stopped after 200 generations. The
transferability function takes as input three behavior descriptors,
computed using the dynamic simulator: (1) the distance covered during
the experiment, (2) the average height of the center of the robot and
(3) the heading of the robot at the end of the experiment. The
similarity measure is the difference between the trajectories in
reality and in simulation (Fig.~\ref{fig:transf}(b)).

We chose a budget of about $10$ evaluations on the real robot
(depending on the treatment). While this number may appear very small,
it is realistic if real experiments are not automated. Additionally,
the problem is simple: only two parameters have to be optimized and
many high-fitness solutions exist.

\begin{figure}[t]
  \begin{center}
    \includegraphics[width=\linewidth]{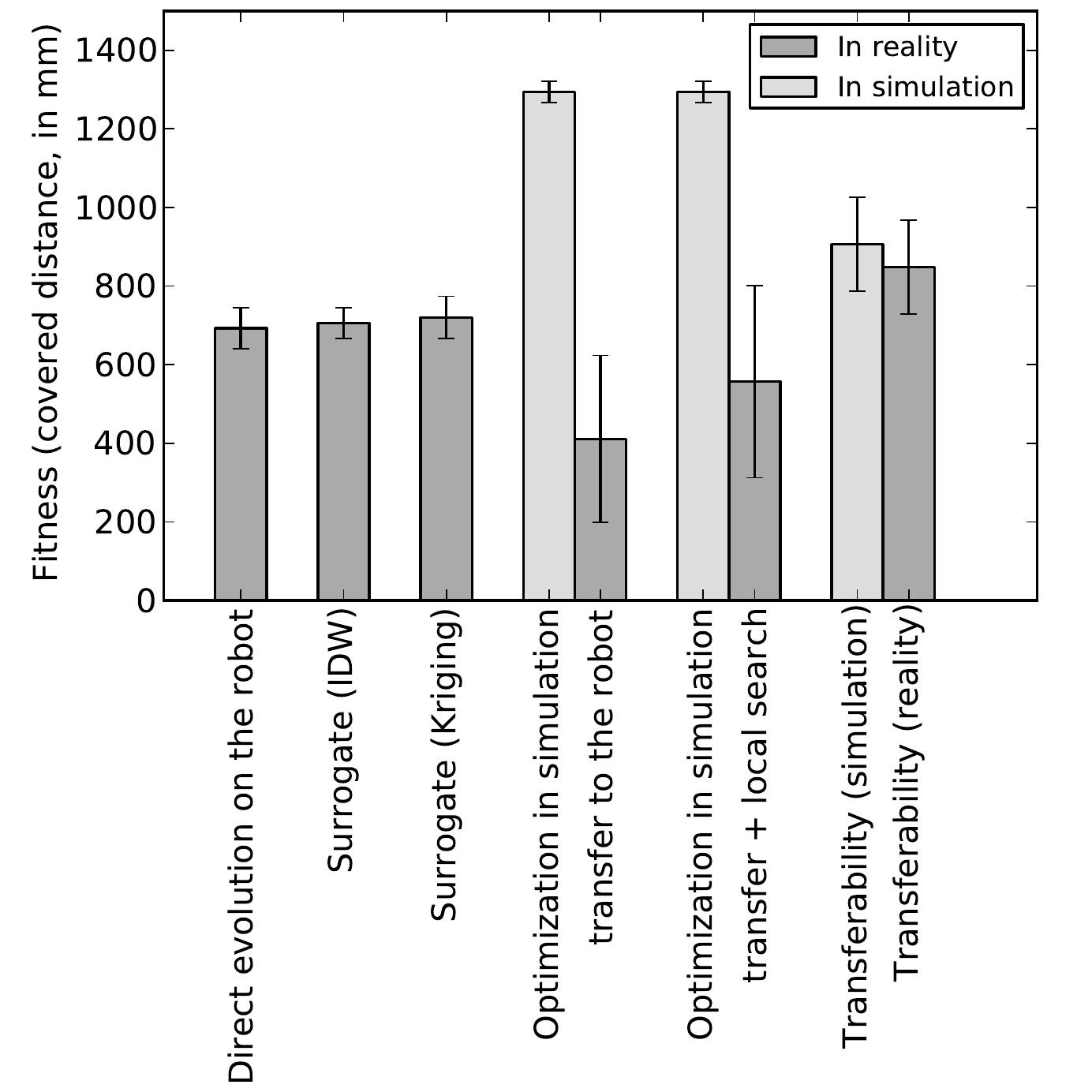}
  \end{center}
  \caption{\label{fig:results}Average distance covered with each of
    the tested treatment (at least $10$ runs for each treatment). The
    transferability approach obtains the best fitness in reality
    (Welsh's t-test, $p \leq 6\cdot10^{-3}$). Error bars indicate one
    unit of standard deviation.}
\end{figure}

We compared the transferability approach to four different treatments,
described belows. 
\paragraph{Direct optimization on the robot.} We used a
population of 4 individuals and 5 generations. This leads to $20$
tests on the real robot. 
\paragraph{Optimization in simulation then transfer to the
    robot.} We expect to observe a reality gap.
\paragraph{Optimization in simulation, transfer to the
    robot and local search.} Parameters of the solutions are modified
using $10$ steps of a stochastic gradient descent, on the real robot.
\paragraph{Surrogate model of the fitness function.} We tested
    IDW~\citep{Shepard1968} and the  Kriging method~\citep{Jin2005b}.

%Each treatment has been repeated at least 10 times to obtain
%statistics.

Results (Fig.\ref{fig:results}) show that solutions found with the
transferability approach have a very similar fitness value in reality
and in simulation, whereas we observe a large reality gap when the
optimization occurs only in simulation. These solutions are also the
ones that work the best on the real robot. It must be emphasized that
the transferability approach did not find the optimal behavior in
simulation (about 1500) nor in reality (about 1500 too). The algorithm
instead found good solutions that work similarly well in simulation
and in reality.

The surrogate models worked better than the optimization in simulation
but it did not significantly improve the result of the direct
optimization on the robot. The addition of a local search stage after
the transfer from simulation to reality significantly improved the
result but final solutions are much worse than those found with the
transferability approach.

We obtained similar results with a second experiment, inspired by
Jakobi's T-maze~\citep{Koos2011}.

\section{Conclusion}
The experimental results validate the relevance of automatically
learning the limits of the simulation to cross the reality gap. The
current implementation relies on several arbitrary choices and many
variants can be designed. More specifically, the choice of the
approximation model and the update strategy need more investigations.

The transferability approach essentially connects a ``slow but
accurate'' evaluation process (the reality) and a second evaluation
process that is ``fast but partially accurate'' (the simulation). The
exact same idea can be used to improve the generalization and the
robustness of optimized controllers in robotics: the reality
corresponds to the evaluation of the controller in many contexts,
whereas the simulation corresponds to its evaluation in a few
contexts. We recently obtained promising results based on this
idea~\citep{Pinville2011}.

Last, we also found that learning the transferability function allows
the design of a fast on-line adaptation algorithm that deports most of
the optimization in a simulation of a self-model~\citep{Koos2011b}.

\footnotesize 
\bibliographystyle{apalike}
\bibliography{biblio}

\end{document}